\documentclass[11pt]{article}
\usepackage{acl2014}
\usepackage{times}
\usepackage{url}
\usepackage{latexsym}
\usepackage[final]{pdfpages} 

\title{Mapping the Economic Crisis: Some Preliminary Investigations}

\author{Pierre Bourreau \\
  CNRS - Lattice - UMR8094\\
  1, rue Maurice Arnoux\\
  92120 Montrouge (France)\\
  {\tt pierre.bourreau@gmail.com}\\\And
  Thierry Poibeau \\
  CNRS - Lattice - UMR8094\\
  1, rue Maurice Arnoux\\
  92120 Montrouge (France)\\
  {\tt thierry.poibeau@ens.fr} \\}

\date{}

\begin{document}
\maketitle
\begin{abstract}
  In this paper we describe our contribution to the PoliInformatics 2014 Challenge on 
  the 2007-2008 financial crisis. We propose a state of the art technique to extract 
  information from texts and provide different representations, giving first a static 
  overview of the domain and then a dynamic representation of its main evolutions. 
  We show that this strategy provides a practical solution to some recent theories 
  in social sciences that are facing a lack of methods and tools to automatically extract 
  information from natural language texts. 
\end{abstract}

\section{Introduction}

This paper describes our contribution to the PoliInformatics 2014 challenge. The organizers of this challenge had made available a series of documents on the 2007-2008 financial crisis. The shared task consisted in developing solutions to address questions such as ``Who was the financial crisis?'' or ``What was the financial crisis?''. Of course, these questions are too complex to receive a simple and direct answer. Consequently our strategy has been to provide tools to process and visualize the most relevant data, so that experts can easily navigate  this flow of information and make sense of the data. While we believe in semi-automatic corpus exploration, we do not think it is possible or even desirable to provide fully automatic answers to the above questions. 

We have mainly used available tools to extract and visualize information. More precisely, we have used the Stanford Named Entity Recognizer \cite{stanford-ner} and the Cortext platform (\url{http://www.cortext.net/}) for information extraction. As for data visualization, we have used Gephi \cite{bastian2009gephi} to observe semantic and social networks, and  the Cortext platform to observe the evolution of the domain over time.  

The rest of the paper is organized as follows: we first present in section 2 the 
corpora selected for this study and in section 3 the application of the Stanford Named Entity Recognizer;
Section~\ref{sec:entities} is dedicated to the first
graph construction procedure based on Gephi; in
Section~\ref{sec:temporal}, the construction of the Sankey diagrams
is explained, before concluding.
Note that all the maps provided 
are also available at \url{http://www.lattice.cnrs.fr/PoliInformatics2014}. 
Some of them integrate interactive facilities (dynamic display of the complete list of terms related to a given topic, etc.) when online.

\section{Corpora Selected for the Analysis}

Different corpora were made available for the unshared task. 
From those, we chose to focus on the reports, and excluded 
bills and auditions as the former seemed to be informationally richer and easier to process. 
Therefore, we worked on
three different files: the Congressional Reports 
\textit{``Wall Street and the Financial Crisis: Anatomy of a Financial
  Collapse''} (referred to as AoC in the rest of the document), 
   \textit{``The Stock Market Plunge: What Happened and
What is Next?''} (referred to as SMP), and the \textit{``Financial
Crisis Inquiry Report''} (referred to as FCIC). Each of these files was
accessible as a PDF, or alternatively as a list of HTML pages,
each page corresponding to a page of the PDF file.

The first task we performed was to transform the HTML files into a
single text document (thanks to a Python
script). In the case of AoC, an option was added to remove page
numbers, which were present in the content part of the HTML
file. Moreover, we normalized the files so that every space break
(line breaks, multiple space breaks or tabular breaks) was changed
into a single space break. After this procedure, we obtained three text
files: AoC.txt, SMP.txt and FCIC.txt.

\section{Named Entity Recognition and Normalization}

The next step was to extract named entities from the different
corpora. In order to do so, we used the Stanford NER, based on
Conditional Random Fields, with MUC tags (Time, Location,
Organization, Person, Money, Percent, Date) \cite{stanford-ner}. We were indeed interested
in extracting organizations and persons so as to get information
about who are the actors of the financial crisis, but dates were also
necessary to be able to represent the temporal evolution of the financial
crisis. 

Because the SMP corpus is too small, we used it for tests but results
are not commented in the present article.

Certain entities appear under different forms. For instance,
``Standard and Poor'' might occur as ``Standard \& Poor'', ``S\&P'' or
``Standard \& Poor's executive board'' (this last sequence in fact refers to a slightly different named entity); in a similar fashion, a person
as ``Mary Schapiro'' may appear as ``Schapiro'', or ``Miss Schapiro'' or
``Chairman Schapiro''. A normalization process is therefore needed.

A large number of references on the subject exist: see, among many others,
 \cite{Zhang:2011mz,Han:2011zr,Gottipati:2011fr,rao11}.
We chose a rather simple approach that performed well on our corpus
based on the following rules
\begin{itemize}
  \item $\mathbf{r_{Org}}$: if two organizations ORG1 and ORG2 are made of
    more than one word, and share the same initials, they are
    considered equal; similarly, if the sequence ORG1 is in ORG2 or ORG2 in ORG1,
    they are considered to refer to the same entity.
  \item $\mathbf{r_{Pers}}$: two persons PERS1 and PERS2 are supposed to
    refer to the same person if the sequence PERS1 is in PERS2 (or PERS2 in PERS1);
    the same hypothesis  stands if the last string in PERS1 (\textit{resp.}~PERS2) is in PERS2
    (\textit{resp.}~PERS1).
\end{itemize}

We did not notice a significant improvement when using more sophisticated methods 
(cf. above) so we keep these simple and traceable set of rules. 

We then had to generalize the approach since the previous strategy is not
enough to cluster together all the relevant linguistic sequences corresponding to a same entity. 
Based on the rules $\mathbf{r_{Org}}$ and $\mathbf{r_{Pers}}$, we tested two 
clustering approaches:

\begin{itemize}
  \item $\mathbf{P_{Max}}$: consider the most frequent entities E1 in
      the set of words S1 and E2 in S2. If they are identical according to
      $\mathbf{r_{ORG}}$, S1 and S2 are merged into a single cluster.
  \item $\mathbf{P_{Av}}$: given \textbf{AV} the average number of occurrences of
    a person or oganization in the corpus, consider the sets A1 $\subseteq$ S1
    and A2 $\subseteq$ S2 made of all entities that occur more than
    \textbf{AV} times in the corpus. If all items E1 $\in$ A1 and all E2
    $\in$ A2 are identical according to $\mathbf{r_{ORG}}$, then S1 and S2 are merged.
\end{itemize}

The normalization subprocess $\mathbf{P_{Max}}$
(\textit{resp.}~$\mathbf{P_{Av}}$) is iterated until a fixpoint is
reached in the set of sets of entities.
After evaluation the normalization strategy based on $\mathbf{P_{Av}}$
turned to be the most efficient one. 
Even if basic, this normalization strategy proved helpful for further analysis. 

\section{Vizualizing entities}\label{sec:entities}

We used the Gephi software \cite{bastian2009gephi} so as to create graphs for each corpus, such
that:
\begin{itemize}
  \item a node corresponds to a cluster of persons or organizations in the
    corresponding corpus (each node is labeled with the most frequent
    entity in the corresponding cluster);
  \item an edge between two nodes corresponds to the number of co-occurrences of the two nodes within the same sentence in the corpus.
\end{itemize}
We chose to consider persons and organizations together since 
they can play a similar role in the event, and metonymy is often used, so that 
a person can refer to a company (and vice versa). 

We then used the Force Atlas algorithm so that only pairs of linked
nodes attract each other. We assign a measure of betweennes centrality
(BC) to each node, which makes nodes with highest degrees (i.e. nodes
with the highest number of links) be bigger; finally, we used a clustering algorithm: the
Louvain Method~\cite{blondel2008fast}, to detect communities in the graph, and we
colored the nodes of each community with a distinct color; intuitively,
we want the number of intra-communities edges to be high, while the number of
inter-communities edges is low. 

\begin{figure*}[h]
\includegraphics[width=0.95\textwidth]{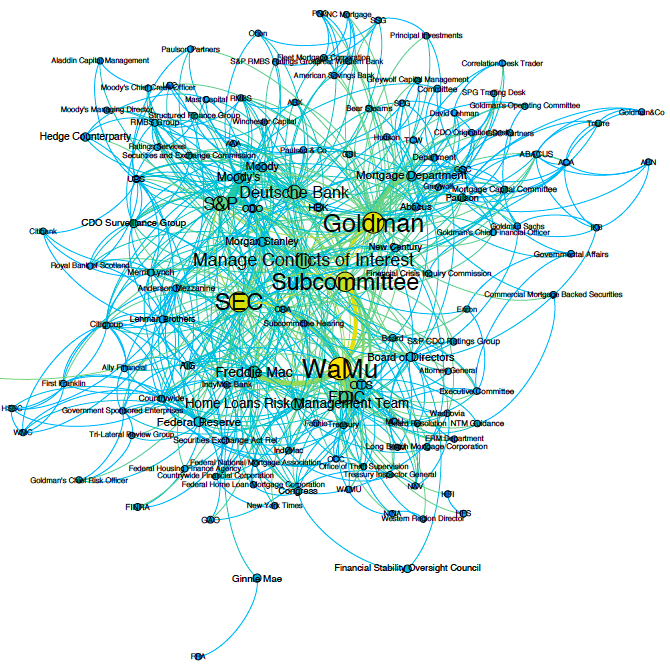} 
\caption{Visualization of links between entities with Gephi (AoC corpus). The original map is dynamic 
and allow end users to navigate the data. 
}
\label{fig1}
\end{figure*}

This process leads to the visualization of two properties: \textit{i)} the
size of a node gives a measure of the dependance of the corresponding set of
entities to other sets of entities. With this measure, we expect
to highlight the role of key organizations and persons during the
financial crisis who have influenced other
organizations/persons, etc.;  \textit{ii)} the same process is then used to highlight strong links between 
 persons and organizations. 

\begin{figure*}[h]
\includegraphics[width=0.95\textwidth]{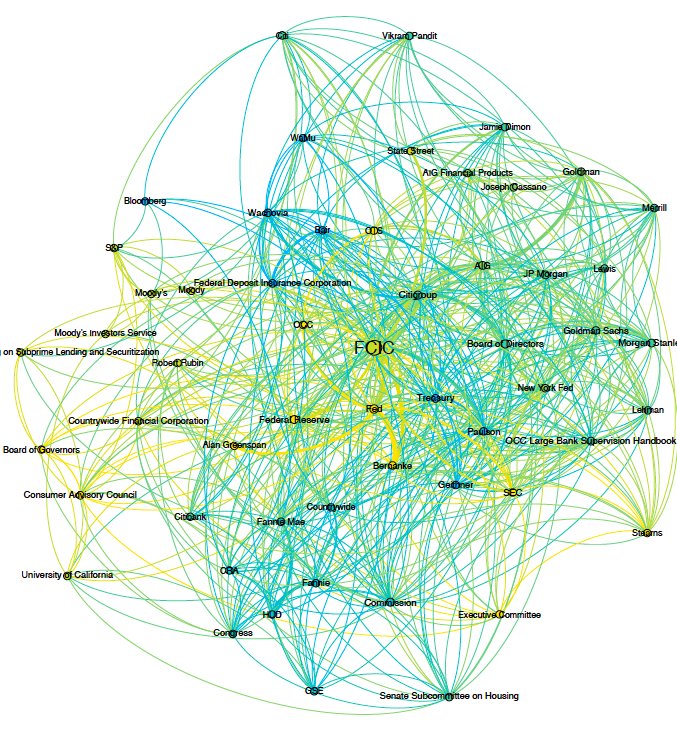}
\caption{Visualization of links between entities with Gephi (FCIC corpus)}
\label{fig2}
\end{figure*}

The results can be seen on figure \ref{fig1} and  figure~\ref{fig2}\footnote{Due to their 
large size and complexity, figures presented in the end of this paper are small and not fully readable; 
however, we hope they give the reader a flavor of the kind of 
representation possible. See \url{http://www.lattice.cnrs.fr/PoliInformatics2014}
to visualize the original maps. }. 
Of course some links are not really informative (for instance the link between an
organization and its CEO); other links express information that may be
registered in different  knowledge bases (see for ex. the link between
\textit{Scott Polakoff}  and   \textit{OTS}, or between
\textit{Fabrice Tourre} and \textit{Goldman Sachs}).   However, we
expect to extract less predictable links that could be of interest
for scholars and experts in the field. 
As an example (and even if this is still not really surprising), we can observe (figure \ref{fig2}) the links between the \textit{Fed 
Consumer Advisory Council} and the \textit{Board of Governors} (for ex. 
\textit{Bernanke, Mark Olson}, and \textit{Kevin Warsh}) since the first group of 
people (\textit{the council}) warns the Board of Governors of the crisis that 
is just beginning.

These visualizations should be explored and tested by specialists who could 
evaluate their real benefits. A historic view on the data would also be 
useful to analyse the dynamics and the evolution of the crisis, through for example
the evolution of terms associated with named entities over different periods of time. 

\section{Visualizing temporal evolution}\label{sec:temporal}

We tried to explore and visualize the temporal evolution of the
financial crisis, more specifically the evolution of the perceived role of organizations over time. 
To do so, we produced Sankey diagrams of the
correlation of organizations and domain related terms in the corpus. 
With this strategy, Sankey diagrams take into account the 
 temporal evolutions of entities and actions along the crisis.

To this aim, we re-used the results of the named entity recognizer (see section 3). 
All the data is extracted from the text itself, including dates. 
A dates is currently  associated only with the sentence it appears in. 
In other words, the scope of dates
(how many events / sentences depend from a given date)
is currently not taken into account but this should be addressed in the near future. 

Organization names as well as dates are normalized, so that only years are kept in the end 
(note however, that a more fine-grained representation, at the month level for example, would be desirable for further investigation). 
Lastly, we only selected
occurrences of years in the range [1990-2020], since documents may refer to past and future legislation 
(but the analysis here is of course not trying to make any prediction for the future). 
Few but some references are made
to the 1929 crisis for instance: those were not relevant for our analysis and were then removed. 

We then run the Lexical Extraction module of the 
Cortext platform (for technical details see
\url{http://docs.cortext.net/lexical-extraction/}) in
order to extract the N most relevant terms in  AoC and
FCIC. We were thus able to extract the most representative 
cooccurrences of named entities along with domain specific terms over time. 

\begin{figure*}[h]
\includegraphics[width=0.95\textwidth]{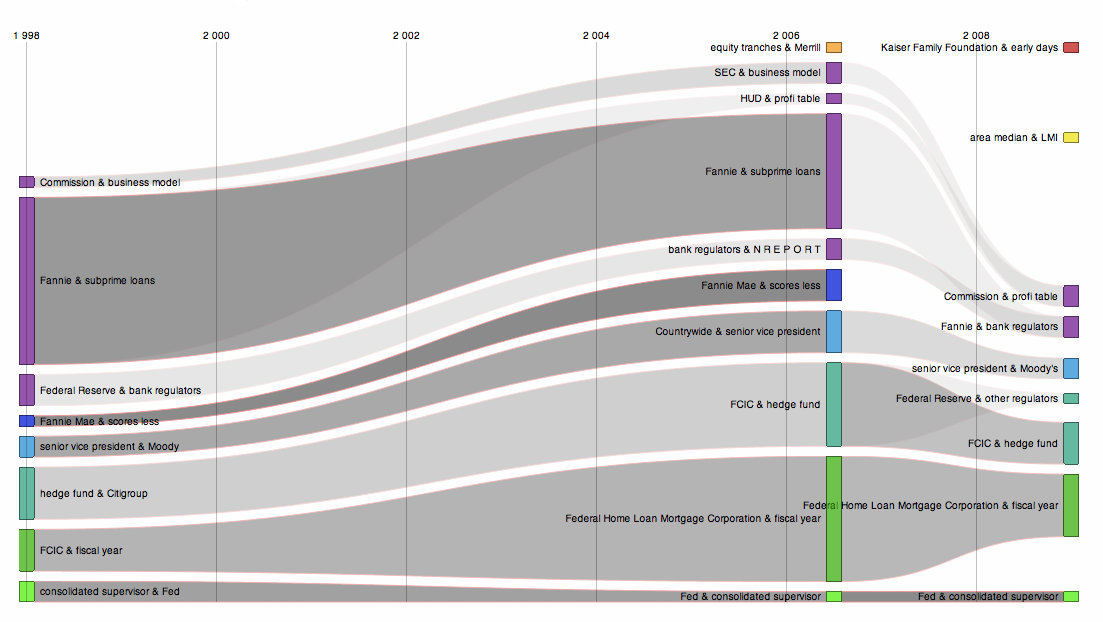}
\caption{Evolution of the links between named entities and topics over time (for FCIC)}
\label{fig3}
\end{figure*}

An overview of the diagram built on FCIC is given on Figure \ref{fig3} for information. 
These representations are in fact interactive: by clicking 
on the different objects (the colored sticks, the grey tubes) it is possible to
get the list of terms that are common (or different) between two different periods of time. 
Associations evolve, hence the tubes can split or merge, depending
on a change in the context, i.e. in the role of an entity or in the way 
it is perceived by other actors.

Figure \ref{fig3} reveals a modification in the data between
2006 and 2008, a period which approximates the start of the financial
crisis. For instance, the stream in purple in this graph reveals many
cooccurrences of Fannie Mae and subprime loans for the period
1990-2007 while for the period 2008-2010, Fannie Mae is 
more closely associated with
'\textit{bank regulators}', or '\textit{Federal Financial Services
  Supervisory Authority}'. In a more general way, all the streams of
data represented in the diagram are dramatically modified after 2007.

\begin{figure*}[h]
\includegraphics[width=0.95\textwidth]{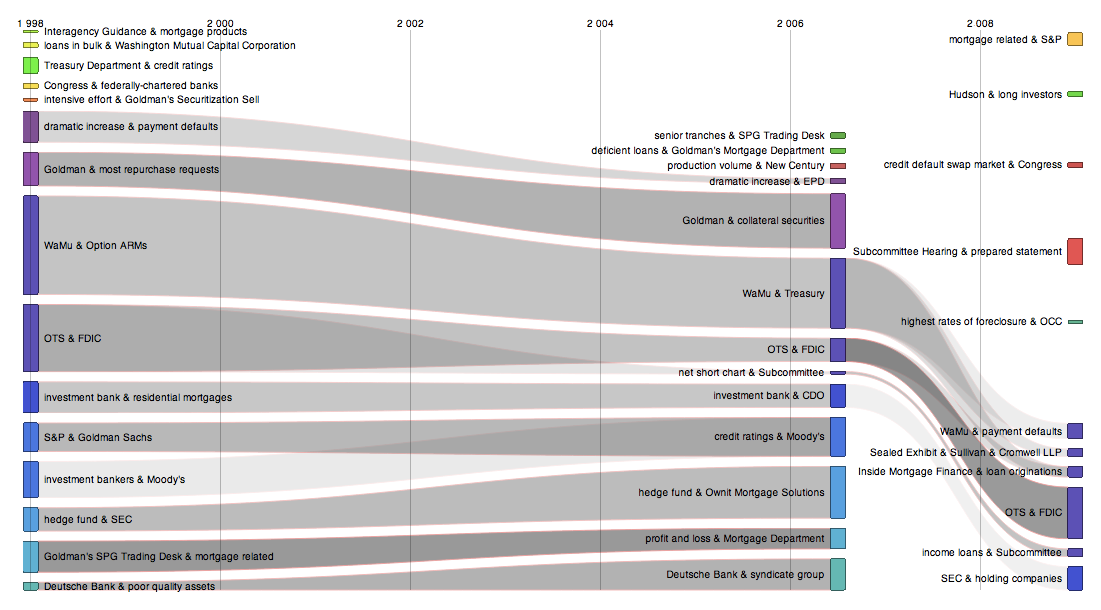} 
\caption{Evolution of the links between named entities and topics over time (for AoC)}
\label{fig4}
\end{figure*}

The AoC diagram (Figure \ref{fig4}) shows the very same modification of data streams,
which suffer from a change after the 2007 time period. 

This representation gives immediate access to key facts. 
For example, the payment default of the  Washington Mutual Bank (WaMu) appears clearly. 
Other examples gives an immediate and meaningful view of the general 
situation over time but a more fine grained analysis would be both possible and desirable.

\hspace{-7mm}

\section{Discussion}

In this paper, we have explored the use of natural language processing tools to extract and visualize 
information on the 2007-2008 financial crisis. 
We first presented a static view of the domain 
based on cooccurrences of named entities (mainly persons and organizations)
within the corpus. 
We then showed how the evolution of the domain can be represented 
taking time into account. In this second experiment we focused on 
co-occurrences of organizations and domain specific terms
and showed that the terms evolve over time as well as their association
with company names, making it possible to get a picture of the domain
without having to read first the whole documentation.  

These representations are inspired by recent theories in social sciences that 
propose not to define a priori groups of interest but to directly take into account actors
and observe regular patterns or regular behaviors \cite{law99,latour05}. Existing natural language processing
tools now make it possible to generate such representation directly from texts, 
which was not possible until recently \cite{venturini}. Traditional representations based 
on co-word analysis are not enough any more and recent advances in named entity 
recognition, relation and event extraction mean that the richness of texts 
can be better exploited.

\section{Future Work}

Perspectives are twofold: on the one hand enhance data analysis so as to provide 
more relevant maps and representations, and on the second hand work closely with 
domain experts and provide interactive ways of navigating the data. 

On the technical side, several improvements could be done. For instance, it would be 
interesting to complete the extraction of technical terms with relevant
verbs so as to obtain more relevant connections between entities. 
The goal would
be to better characterize the role played by the different entities. Also, dates
considered in the temporal analysis could be scaled in months or
days. In the same direction, it could be interesting to test different
named entities normalization strategies, and to test our methods on
more texts from different sources.

Concerning interactions with experts, it is clear that end users 
could provide a very valuable contribution in the selection of relevant data
as well as in the way they are linked and mapped. 
Some experiments are currently being done with a focus group gathering social science
as well as information science experts. 
They will  assess that the solution 
is useful and workable and more importantly, will give feedback
so as to provide better solutions. 

\section*{Acknowledgments}

The authors are thankful to Elisa Omodei for her help in using the
Cortext platform. We would also like to thank the two reviewers
from the PoliInformatics challenge for their useful feedback and
 suggestions on the work done. We are also grateful to the 
 experts involved in our focus group. 
 
This research was partially funded by the 
CNRS Mastodons Aresos project. 

\bibliographystyle{acl}
\bibliography{Biblio}

\end{document}